\documentclass[11pt,a4paper]{article}
\usepackage[hyperref]{acl2019}
\usepackage{times}
\usepackage{latexsym}
\usepackage{graphicx}
\usepackage{algorithm}
\usepackage{algorithmicx}
\usepackage{algpseudocode}
\usepackage{multicol}
\usepackage{multirow}
\usepackage{array}
\usepackage{booktabs}
\usepackage{amsmath}
\usepackage{longtable}
\usepackage{float}

\usepackage{url}
\usepackage{ragged2e}
\usepackage{tabularx}
\usepackage{adjustbox}
\usepackage{amssymb}
\usepackage[T1]{fontenc}

\aclfinalcopy 
\def\aclpaperid{1206} 

\makeatletter
\newcommand{\printfnsymbol}[1]{%
  \textsuperscript{\@fnsymbol{#1}}%
}
\makeatother

\title{Semantic Parsing with Dual Learning}

\author{Ruisheng Cao\thanks{\ \ Ruisheng Cao and Su Zhu are co-first authors and contribute equally to this work.}, Su Zhu\printfnsymbol{1}, Chen Liu, Jieyu Li \and Kai Yu  \thanks{\ \ The corresponding author is Kai Yu.}\\
  MoE Key Lab of Artificial Intelligence\\
  SpeechLab, Department of Computer Science and Engineering\\
  Shanghai Jiao Tong University, Shanghai, China\\
  {\tt \{211314,paul2204,chris-chen,oracion,kai.yu\}@sjtu.edu.cn} \\}

\date{}

\begin{document}
\maketitle

\begin{abstract}
Semantic parsing converts natural language queries into structured logical forms. The paucity of annotated training samples is a fundamental challenge in this field. In this work, we develop a semantic parsing framework with the dual learning algorithm, which enables a semantic parser to make full use of data (labeled and even unlabeled) through a dual-learning game. This game between a primal model (semantic parsing) and a dual model (logical form to query) forces them to regularize each other, and can achieve feedback signals from some prior-knowledge. By utilizing the prior-knowledge of logical form structures, we propose a novel reward signal at the surface and semantic levels which tends to generate complete and reasonable logical forms. Experimental results show that our approach achieves new state-of-the-art performance on ATIS dataset and gets competitive performance on \textsc{Overnight} dataset.
\end{abstract}

\section{Introduction}
Semantic parsing is the task of mapping a natural language query into a logical form \cite{zelle1996learning,wong2007learning,zettlemoyer2007online,lu2008generative,zettlemoyer2012learning}. A logical form is one type of meaning representation understood by computers, which usually can be executed by an executor to obtain the answers. 



The successful application of recurrent neural networks (RNN) in a variety of NLP tasks \cite{bahdanau2014neural,sutskever2014sequence,vinyals2015grammar} has provided strong impetus to treat semantic parsing as a sequence-to-sequence (Seq2seq) problem \cite{jia2016data,dong2016language}. This approach generates a logical form based on the input query in an end-to-end manner but still leaves two main issues: (1) lack of labeled data and (2) constrained decoding. 

Firstly, semantic parsing relies on sufficient labeled data. However, data annotation of semantic parsing is a labor-intensive and time-consuming task. Especially, the logical form is unfriendly for human annotation.

Secondly, unlike natural language sentences, a logical form is strictly structured. For example, the lambda expression of ``show flight from ci0 to ci1'' is {\tt ( lambda \$0 e ( and ( from \$0 ci0 ) ( to \$0 ci1 ) ( flight \$0 ) ) )}. If we make no constraint on decoding, the generated logical form may be invalid or incomplete at \emph{surface} and \emph{semantic} levels.
\begin{description}
\item[Surface] The generated sequence should be structured as a complete logical form. For example, left and right parentheses should be matched to force the generated sequence to be a valid tree.
\item[Semantic] Although the generated sequence is a legal logical form at surface level, it may be meaningless or semantically ill-formed. For example, the predefined binary predicate \verb|from| takes no more than two arguments. The first argument must represent a \verb|flight| and the second argument should be a \verb|city|. 
\end{description}
To avoid producing incomplete or semantically ill-formed logical forms, the output space must be constrained.

\begin{figure*}[htbp]
    \centering
    \includegraphics[width=1\textwidth]{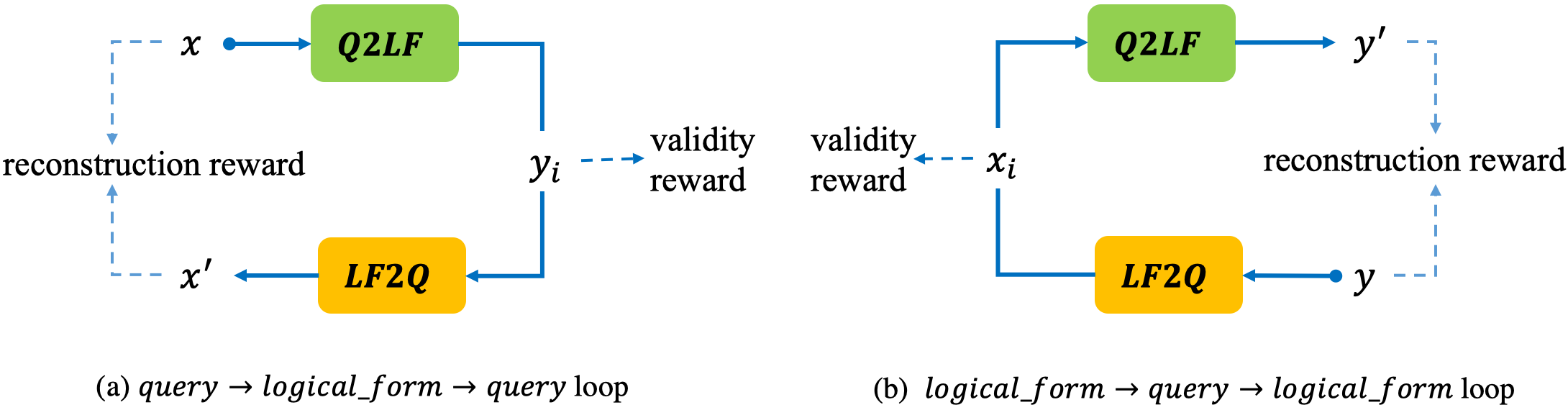}
    \caption[width=1.0\textwidth]{An overview of dual semantic parsing framework. The primal model ($Q2LF$) and dual model ($LF2Q$) can form a closed cycle. But there are two different directed loops, depending on whether they start from a query or logical form. Validity reward is used to estimate the quality of the middle generation output, and reconstruction reward is exploited to avoid information loss. The primal and dual models can be pre-trained and fine-tuned with labeled data to keep the models effective.}
    \label{fig:dual_learning}
\end{figure*}

In this paper, we introduce a semantic parsing framework (see Figure \ref{fig:dual_learning}) by incorporating dual learning \cite{he2016dual} to tackle the problems mentioned above. In this framework, we have a primal task (query to logical form) and a dual task (logical form to query). They can form a closed loop, and generate informative feedback signals to train the primal and dual models even without supervision. In this loop, the primal and dual models restrict or regularize each other by generating intermediate output in one model and then checking it in the other. Actually, it can be viewed as a method of data augmentation. Thus it can leverage unlabeled data (queries or synthesized logical forms) in a more effective way which helps alleviate the problem of lack of annotated data.

In the dual learning framework, the primal and dual models are represented as two agents and they teach each other through a reinforcement learning process. To force the generated logical form complete and well-formed, we newly propose a \emph{validity reward} by checking the output of the primal model at the surface and semantic levels.

We evaluate our approach on two standard datasets: ATIS and \textsc{Overnight}. The results show that our method can obtain significant improvements over strong baselines on both datasets with fully labeled data, and even outperforms state-of-the-art results on ATIS. With additional logical forms synthesized from rules or templates, our method is competitive with state-of-the-art systems on \textsc{Overnight}. Furthermore, our method is compatible with various semantic parsing models. We also conduct extensive experiments to further investigate our framework in semi-supervised settings, trying to figure out why it works.


The main contributions of this paper are summarized as follows:
\begin{itemize}
    \item An innovative semantic parsing framework based on dual learning is introduced, which can fully exploit data (labeled or unlabeled) and incorporate various prior-knowledge as feedback signals. We are the first to introduce dual learning in semantic parsing to the best of our knowledge. 
    \item We further propose a novel validity reward focusing on the surface and semantics of logical forms, which is a feedback signal indicating whether the generated logical form is well-formed. It involves the prior-knowledge about structures of logical forms predefined in a domain.
    
    \item We conduct extensive experiments on ATIS and \textsc{Overnight} benchmarks. The results show that our method achieves new state-of-the-art performance (test accuracy 89.1\%) on ATIS dataset and gets competitive performance on \textsc{Overnight} dataset.
\end{itemize}


\section{Primal and Dual Tasks of Semantic Parsing}
Before discussing the dual learning algorithm for semantic parsing, we first present the primal and dual tasks (as mentioned before) in detail. The primal and dual tasks are modeled on the attention-based Encoder-Decoder architectures (i.e. Seq2seq) which have been successfully applied in neural semantic parsing \cite{jia2016data,dong2016language}. We also include copy mechanism \cite{gulcehre2016pointing,see2017get} to tackle unknown tokens.

\subsection{Primal Task}
The primal task is semantic parsing which converts queries into logical forms ($Q2LF$). Let $x=x_1 \cdots x_{|x|}$ denote the query, and $y=y_1 \cdots y_{|y|}$ denote the logical form. An \emph{encoder} is exploited to encode the query $x$ into vector representations, and a \emph{decoder} learns to generate logical form $y$ depending on the encoding vectors.
\vspace{0.2em}

\noindent\textbf{Encoder}\quad Each word $x_i$ is mapped to a fixed-dimensional vector by a word embedding function $\psi(\cdot)$ and then fed into a bidirectional LSTM \cite{hochreiter1997long}. The hidden vectors are recursively computed at the $i$-th time step via:
\begin{align}
\overrightarrow{\textbf{h}_i}=&\text{f}_\text{{LSTM}}(\psi(x_i), \overrightarrow{\textbf{h}}_{i-1}), i=1,\cdots,|x|\\
\overleftarrow{\textbf{h}_i}=&\text{f}_\text{{LSTM}}(\psi(x_i), \overleftarrow{\textbf{h}}_{i+1}), i=|x|,\cdots,1\\
\textbf{h}_i=&[\overrightarrow{\textbf{h}}_i;\overleftarrow{\textbf{h}}_i]
\end{align}
where $[\cdot;\cdot]$ denotes vector concatenation, $\textbf{h}_i\in \mathbb{R}^{2n}$, $n$ is the number of hidden cells and $\text{f}_\text{{LSTM}}$ is the LSTM function.
\vspace{0.2em}

\noindent\textbf{Decoder}\quad Decoder is an unidirectional LSTM with the attention mechanism \cite{luong2015effective}. The hidden vector at the $t$-th time step is computed by $\textbf{s}_t=\text{f}_\text{{LSTM}}(\phi(y_{t-1}),\textbf{s}_{t-1})$, where $\phi(\cdot)$ is the token embedding function for logical forms and $\textbf{s}_t\in \mathbb{R}^n$. The hidden vector of the first time step is initialized as $\textbf{s}_0=\overleftarrow{\textbf{h}}_1$. The attention weight for the current step $t$ of the decoder, with the $i$-th step in the encoder is $a_i^t=\frac{\text{exp}(u_i^t)}{\sum_{j=1}^{|x|}\text{exp}(u_j^t)}$ and 
\begin{align}
u_i^t=&\textbf{v}^T  \text{tanh}(\textbf{W}_1\textbf{h}_i+\textbf{W}_2\textbf{s}_t+\textbf{b}_{a})
\end{align}
where $\textbf{v}, \textbf{b}_{a}\in \mathbb{R}^n$, and $\textbf{W}_1 \in \mathbb{R}^{n\times 2n},\textbf{W}_2 \in \mathbb{R}^{n\times n}$ are parameters. Then we compute the vocabulary distribution $P_{gen}(y_t|y_{<t},x)$ by
\begin{align}
\textbf{c}_t=&\sum_{i=1}^{|x|}a_i^t \textbf{h}_i\\
P_{gen}(y_t|y_{<t},x)=&\text{softmax}(\textbf{W}_o[\textbf{s}_t;\textbf{c}_t]+\textbf{b}_o)
\end{align}
where $\textbf{W}_o \in \mathbb{R}^{|\mathcal{V}_y|\times 3n}$, $\textbf{b}_{o}\in \mathbb{R}^{|\mathcal{V}_y|}$ and $|\mathcal{V}_y|$ is the output vocabulary size. Generation ends once an end-of-sequence token ``EOS'' is emitted.
\vspace{0.2em}

\noindent\textbf{Copy Mechanism}\quad We also include copy mechanism to improve model generalization following the implementation of \citet{see2017get}, a hybrid between \citet{nallapati2016abstractive} and pointer network \cite{gulcehre2016pointing}. The predicted token is from either a fixed output vocabulary $\mathcal{V}_y$ or raw input words $x$. We use sigmoid gate function $\sigma$ to make a soft decision between generation and copy at each step $t$.
\begin{align}
g_t = &\sigma(\textbf{v}^T_g[\textbf{s}_t;\textbf{c}_t;\phi(y_{t-1})]+b_g)\\
\begin{split}
P(y_t|y_{<t},x) = & g_t P_{gen}(y_t|y_{<t},x)\\
& +(1-g_t) P_{copy}(y_t|y_{<t},x) \label{eq:ptr}
\end{split}
\end{align}
where $g_t\in[0,1]$ is the balance score, $\textbf{v}_g$ is a weight vector and $b_g$ is a scalar bias. Distribution $P_{copy}(y_t|y_{<t},x)$ will be described as follows.
\vspace{0.2em}

\noindent\textbf{Entity Mapping}\quad Although the copy mechanism can deal with unknown words, many raw words can not be directly copied to be part of a logical form. For example, {\tt kobe bryant} is represented as {\tt en.player.kobe\_bryant} in \textsc{Overnight} \cite{wang2015building}. It is common that entities are identified by Uniform Resource Identifier (URI, \citealp{klyne2006resource}) in a knowledge base. Thus, a mapping from raw words to URI is included after copying. Mathematically, $P_{copy}$ in Eq.\ref{eq:ptr} is calculated by:
\begin{align*}
P_{copy}(y_t=w|y_{<t},x)=&\sum_{i,j:\ KB(x_{i:j})=w}\sum_{k=i}^ja_k^t
\end{align*}
where $i<j$, $a_k^t$ is the attention weight of position $k$ at decoding step $t$, $KB(\cdot)$ is a dictionary-like function mapping a specific noun phrase to the corresponding entity token in the vocabulary of logical forms.

\subsection{Dual Model}
The dual task ($LF2Q$) is an inverse of the primal task, which aims to generate a natural language query given a logical form. We can also exploit the attention-based Encoder-Decoder architecture (with copy mechanism or not) to build the dual model. 

\noindent\textbf{Reverse Entity Mapping}\quad Different with the primal task, we reversely map every possible KB entity $y_t$ of a logical form to the corresponding noun phrase before query generation, $KB^{-1}(y_t)$\footnote{$KB^{-1}(\cdot)$ is the inverse operation of $KB(\cdot)$, which returns the set of all corresponding noun phrases given a KB entity.}. Since each KB entity may have multiple aliases in the real world, e.g. {\tt kobe bryant} has a nickname {\tt the black mamba}, we make different selections in two cases:

\begin{itemize}
    \item For paired data, we select the noun phrase from $KB^{-1}(y_t)$, which exists in the query.
    \item For unpaired data, we randomly select one from $KB^{-1}(y_t)$.
\end{itemize}



\section{Dual learning for Semantic Parsing}

In this section, we present a semantic parsing framework with dual learning. We use one agent to represent the model of the primal task ($Q2LF$) and another agent to represent the model of the dual task ($LF2Q$), then design a two-agent game in a closed loop which can provide quality feedback to the primal and dual models even if only queries or logical forms are available. As the feedback reward may be non-differentiable, reinforcement learning algorithm \cite{sutton2018reinforcement} based on policy gradient \cite{sutton2000policy} is applied for optimization.

Two agents, $Q2LF$ and $LF2Q$, participate in the collaborative game with two directed loops as illustrated in Figure \ref{fig:dual_learning}. One loop \verb|query->logical_form->query| starts from a query, generates possible logical forms by agent $Q2LF$ and tries to reconstruct the original query by $LF2Q$. The other loop \verb|logical_form->query->logical_form| starts from the opposite side. Each agent will obtain quality feedback depending on reward functions defined in the directed loops.


\subsection{Learning algorithm}
\label{sec:learning_algorithm}
Suppose we have fully labeled dataset $\mathcal{T}=\{\left<x,y\right>\}$, unlabeled dataset $\mathcal{Q}$ with only queries if available, and unlabeled dataset $\mathcal{LF}$ with only logical forms if available. We firstly pre-train the primal model $Q2LF$ and the dual model $LF2Q$ on $\mathcal{T}$ by maximum likelihood estimation (MLE). Let $\Theta_{Q2LF}$ and $\Theta_{LF2Q}$ denote all the parameters of $Q2LF$ and $LF2Q$ respectively. Our learning algorithm in each iteration consists of three parts:
\subsubsection{Loop starts from a query}
As shown in Figure \ref{fig:dual_learning} (a), we sample a query $x$ from $\mathcal{Q}\cup\mathcal{T}$ randomly. Given $x$, $Q2LF$ model could generate $k$ possible logical forms ${y}_1,{y}_2,\cdots,{y_k}$ via beam search ($k$ is beam size). For each ${y}_i$, we can obtain a validity reward $R_q^{val}(y_i)$ (a scalar) computed by a specific reward function which will be discussed in Section \ref{sec:validity reward}. After feeding ${y}_i$ into $LF2Q$, we finally get a reconstruction reward $R_q^{rec}(x, y_i)$ which forces the generated query as similar to $x$ as possible and will be discussed in Section \ref{sec:rec_reward}. 

A hyper-parameter $\alpha$ is exploited to balance these two rewards in $r^q_i=\alpha R_q^{val}(y_i)+(1-\alpha)R_q^{rec}(x, y_i)$, where $\alpha \in [0, 1]$.

By utilizing policy gradient \cite{sutton2000policy}, the stochastic gradients of $\Theta_{Q2LF}$ and $\Theta_{LF2Q}$ are computed as:

\vspace{-1em}
{\small
\begin{align*}
\nabla_{\Theta_{Q2LF}}\hat{E}[r]=&\frac{1}{k}\sum_{i=1}^k r^q_i\nabla_{\Theta_{Q2LF}}\log P(y_i|x;\Theta_{Q2LF})\\
\nabla_{\Theta_{LF2Q}}\hat{E}[r]=&\frac{1-\alpha}{k}\sum_{i=1}^k\nabla_{\Theta_{LF2Q}}\log P(x|y_i;\Theta_{LF2Q})
\end{align*}
}
\vspace{-1.5em}
\subsubsection{Loop starts from a logical form}
As shown in Figure \ref{fig:dual_learning} (b), we sample a logical form $y$ from $\mathcal{LF}\cup\mathcal{T}$ randomly. Given $y$, $LF2Q$ model generates $k$ possible queries ${x}_1,{x}_2,\cdots,{x_k}$ via beam search. For each ${x}_i$, we can obtain a validity reward $R_{lf}^{val}(x_i)$ (a scalar)  which will also be discussed in Section \ref{sec:validity reward}. After feeding ${x}_i$ into $Q2LF$, we can also get a reconstruction reward $R_{lf}^{rec}(y, x_i)$, which forces the generated logical form as similar to $y$ as possible and will be discussed in Section \ref{sec:rec_reward}. 

A hyper-parameter $\beta$ is also exploited to balance these two rewards by $r^{lf}_i=\beta R_{lf}^{val}(x_i)+(1-\beta)R_{lf}^{rec}(y, x_i)$, where $\beta \in [0, 1]$.

By utilizing policy gradient, the stochastic gradients of $\Theta_{Q2LF}$ and $\Theta_{LF2Q}$ are computed as:

\vspace{-1em}
{\small
\begin{align*}
\nabla_{\Theta_{LF2Q}}\hat{E}[r]=&\frac{1}{k}\sum_{i=1}^k r^{lf}_i\nabla_{\Theta_{LF2Q}}\log P(x_i|y;\Theta_{LF2Q})\\
\nabla_{\Theta_{Q2LF}}\hat{E}[r]=&\frac{1-\beta}{k}\sum_{i=1}^k\nabla_{\Theta_{Q2LF}}\log P(y|x_i;\Theta_{Q2LF})
\end{align*}
}
\vspace{-1em}
\subsubsection{Supervisor guidance}
The previous two stages are unsupervised learning processes, which need no labeled data. If there is no supervision for the primal and dual models after pre-training, these two models would be rotten especially when $\mathcal{T}$ is limited.

To keep the learning process stable and prevent the models from crashes, we randomly select samples from $\mathcal{T}$ to fine-tune the primal and dual models by maximum likelihood estimation (MLE). Details about the dual learning algorithm for semantic parsing are provided in Appendix \ref{app:dual_alg}.


\subsection{Reward design}
As mentioned in Section \ref{sec:learning_algorithm}, there are two types of reward functions in each loop: validity reward ($R_q^{val}$, $R_{lf}^{val}$) and reconstruction reward ($R_q^{rec}$, $R_{lf}^{rec}$). But each type of reward function may be different in different loops. 

\subsubsection{Validity reward}
\label{sec:validity reward}
Validity reward is used to evaluate the quality of intermediate outputs in a loop (see Figure \ref{fig:dual_learning}). In the loop starts from a query, it indicates whether the generated logical forms are well-formed at the surface and semantic levels. While in the loop starts from a logical form, it indicates how natural and fluent the intermediate queries are.

\noindent\textbf{Loop starts from a query:} We estimate the quality of the generated logical forms at two levels, i.e. surface and semantics. Firstly, we check whether the logical form is a complete tree without parentheses mismatching. As for semantics, we check whether the logical form is understandable without errors like type inconsistency. It can be formulated as 
\begin{align}\label{eq:val}
R_q^{val}(y)=\text{grammar\_error\_indicator}(y)
\end{align}
which returns $1$ when $y$ has no error at the surface and semantic levels, and returns $0$ otherwise.

If there exists an executing program or search engine for logical form $y$, e.g. dataset \textsc{Overnight} \cite{wang2015building}, $\text{grammar\_error\_indicator}(\cdot)$ has been included.


Otherwise, we should construct a grammar error indicator based on the ontology of the corresponding dataset. For example, a specification of ATIS can be extracted by clarifying all (1) entities paired with corresponding types, (2) unary/binary predicates with argument constraints (see Table \ref{tab:ontology}). Accordingly, Algorithm \ref{alg:evaluator} abstracts the procedure of checking the surface and semantics for a logical form candidate $y$ based on the specification.
\begin{table}[htbp]
\centering
\begin{adjustbox}{width=7.7cm}
\begin{tabular}{|c|c|c|c|}
\hline
\textbf{category} & \textbf{type} & \multicolumn{2}{c|}{\textbf{instances}} \\
\hline
\multirow{3}*{entity} & me & \multicolumn{2}{c|}{lunch:me, snack:me} \\
\cline{2-4} & al & \multicolumn{2}{c|}{delta:al, usair:al} \\
\cline{2-4} & pd & \multicolumn{2}{c|}{morning:pd, late:pd} \\
\hline
\textbf{category} & \textbf{instances} & \multicolumn{2}{c|}{\textbf{args0}} \\
\hline
\multirow{3}*{unary} & \_city & \multicolumn{2}{c|}{ci} \\
\cline{2-4} & \_airport & \multicolumn{2}{c|}{ap} \\
\cline{2-4} & \_oneway & \multicolumn{2}{c|}{flight} \\
\hline
\textbf{category} & \textbf{instances} & \textbf{\ \ \ args0\ \ } & \textbf{args1} \\
\hline
\multirow{3}*{binary} & \_from & flight & ci \\
\cline{2-4} & \_services & al & ci \\
\cline{2-4} & \_during\_day & flight & pd \\
\hline
\end{tabular}
\end{adjustbox}
\caption{A truncated specification for ATIS.}
\label{tab:ontology}%
\end{table}%

\begin{algorithm}[htp]
\caption{Grammar error indicator on ATIS}
\label{alg:evaluator}
\begin{algorithmic}[1]
\Require Logical form string $y$; specification $\mathcal{D}$
\Ensure 1/0, whether $y$ is valid
\If{$to\_lisp\_tree(y)$ succeed}
\State{$lispTree\leftarrow to\_lisp\_tree(y)$}

\Comment{using Depth-First-Search for $lispTree$ }
\If{$type\_ consistent(lispTree,\mathcal{D})$}
\State{\Return 1}
\EndIf
\EndIf
\State{\Return 0}
\end{algorithmic}
\end{algorithm}

\noindent\textbf{Loop starts from a logical form:} A language model (LM) is exploited to evaluate the quality of intermediate queries \cite{mikolov2010recurrent}. We apply length-normalization \cite{wu2016google} to make a fair competition between short and long queries.
\begin{align}\label{eq:lm}
R_{lf}^{val}(x)=&\log LM_q(x)/Length(x),
\end{align}
where $LM_q(\cdot)$ is a language model pre-trained on all the queries of $\mathcal{Q}\cup\mathcal{T}$ (referred in Section \ref{sec:learning_algorithm}). 

\subsubsection{Reconstruction reward}
\label{sec:rec_reward}
Reconstruction reward is used to estimate how similar the output of one loop is compared with the input. We take log likelihood as reconstruction rewards for the loop starts from a query and the loop starts from a logical form. Thus,
\begin{align*}
R_q^{rec}(x, y_{i})=&\log P(x|{y}_{i};\Theta_{LF2Q}) \\
R_{lf}^{rec}(y, x_{i})=&\log P(y|{x}_{i};\Theta_{Q2LF})
\end{align*}
where $y_{i}$ and $x_{i}$ are intermediate outputs.

\section{Experiment}
In this section, we evaluate our framework on the ATIS and \textsc{Overnight} datasets. 

\subsection{Dataset}
\noindent\textbf{ATIS} We use the preprocessed version provided by \citet{dong2018coarse}, where natural language queries are lowercased and stemmed with NLTK \cite{loper2002nltk}, and entity mentions are replaced by numbered markers. We also leverage an external lexicon that maps word phrases (e.g., {\tt first class}) to entities (e.g., {\tt first:cl}) like what \citet{jia2016data} did. 

\noindent\textbf{\textsc{Overnight}} It contains natural language paraphrases paired with logical forms across eight domains. We follow the traditional 80\%/20\% train/valid splits as \citet{wang2015building} to choose the best model during training. 

ATIS and \textsc{Overnight} never provide unlabeled queries. To test our method in semi-supervised learning, we keep a part of the training set as fully labeled data and leave the rest as unpaired queries and logical forms which simulate unlabeled data. 

\subsection{Synthesis of logical forms}

Although there is no unlabeled query provided in most semantic parsing benchmarks, it should be easy to synthesize logical forms. Since a logical form is strictly structured and can be modified from the existing one or created from simple grammars, it is much cheaper than query collection. Our synthesized logical forms are public \footnote{\url{https://github.com/RhythmCao/Synthesized-Logical-Forms}}.

\subsubsection{Modification based on ontology}
\begin{figure}[htbp]
    \centering
    \includegraphics[width=7.7cm]{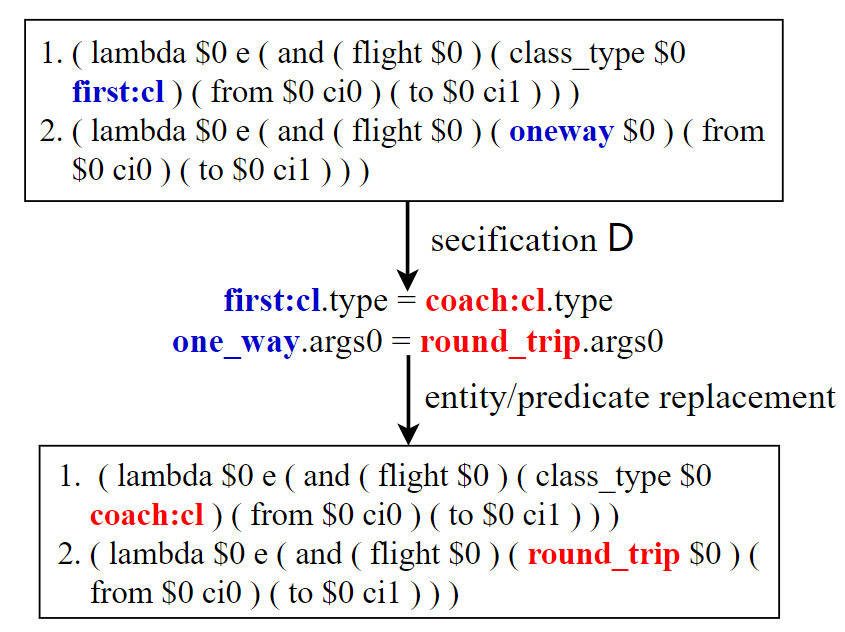}
    \caption[width=0.5\textwidth]{Synthesis of logical forms by replacement.}
    \label{fig:aug_ontology}
\end{figure}

On ATIS, we randomly sample a logical form from the training set, and select one entity or predicate for replacement according to the specification in Table \ref{tab:ontology}. If the new logical form after replacement is valid and never seen, it is added to the unsupervised set. 4592 new logical forms are created for ATIS.
An example is shown in Figure \ref{fig:aug_ontology}.
\vspace{-0.8em}

\vspace{-0.8em}

\subsubsection{Generation based on grammar}
\citet{wang2015building} proposed an underlying grammar to generate logical forms along with their corresponding canonical utterances on \textsc{Overnight}, which can be found in SEMPRE \footnote{https://github.com/percyliang/sempre}. We reorder the entity instances (e.g., \textsc{EntityNP}) of one type (e.g., \textsc{TypeNP}) in grammar files to generate new logical forms. We could include new entity instances if we want more unseen logical forms, but we didn't do that actually. Finally, we get about 500 new logical forms for each domain on average. More examples can be found in Appendix \ref{app:examples_of_logical_form}.
 

\subsection{Experimental settings}
\subsubsection{Base models}
We use $200$ hidden units and $100$-dimensional word vectors for all encoders and decoders of $Q2LF$ and $LF2Q$ models. The LSTMs used are in single-layer. Word embeddings on query side are initialized by Glove6B \cite{pennington2014glove}. Out-of-vocabulary words are replaced with a special token $\left<unk\right>$. Other parameters are initialized by uniformly sampling within the interval $[-0.2, 0.2]$. The language model we used is also a single-layer LSTM with $200$ hidden units and $100$-dim word embedding layer.

\subsubsection{Training and decoding}
We individually pre-train $Q2LF$/$LF2Q$ models using only labeled data and language model $LM_q$ using both labeled and unlabeled queries. The language model is fixed for calculating reward. The hyper-parameters $\alpha$ and $\beta$ are selected according to validation set ($0.5$ is used), and beam size $k$ is selected from $\{3,5\}$. The batch size is selected from $\{10,20\}$. We use optimizer Adam \cite{kingma2014adam} with learning rate $0.001$ for all experiments. Finally, we evaluate the primal model ($Q2LF$, semantic parsing) and report test accuracy on each dataset.


\subsection{Results and analysis}
We perform a \textsc{Pseudo} baseline following the setup in \citet{sennrich2015improving} and \citet{guo2018question}. The pre-trained $LF2Q$ or $Q2LF$ model is used to generate pseudo $\left<query, logical\ form\right>$ pairs from unlabeled logical forms or unlabeled queries, which extends the training set. The pseudo-labeled data is used carefully with a discount factor (0.5) in loss function \cite{lee2013pseudo}, when we train $Q2LF$ by supervised training.

\subsubsection{Main results}
\begin{table*}[htbp]
  \centering{
    \small
    
\begin{adjustbox}{width=\textwidth}
    \begin{tabular}{p{19em}|cccccccc|c}
    \hline
    \textbf{Method} & \multicolumn{1}{p{2.0em}}{\textbf{Bas.}} & \multicolumn{1}{p{2.0em}}{\textbf{Blo.}} & \multicolumn{1}{p{2.0em}}{\textbf{Cal.}} & \multicolumn{1}{p{2.0em}}{\textbf{Hou.}} & \multicolumn{1}{p{2.0em}}{\textbf{Pub.}} & \multicolumn{1}{p{2.0em}}{\textbf{Rec.}} & \multicolumn{1}{p{2.0em}}{\textbf{Res.}} & \multicolumn{1}{p{2.0em}|}{\textbf{Soc.}} & \multicolumn{1}{p{2.0em}}{\textbf{Avg.}} \\
    \hline
    SPO \cite{wang2015building} & 46.3  & 41.9  & 74.4  & 54.0  & 59.0  & 70.8  & 75.9  & 48.2  & 58.8  \\
    DSP-C \cite{xiao2016sequence} & 80.5  & 55.6  & 75.0  & 61.9  & 75.8  & \_ & 80.1  & 80.0  & 72.7  \\
    \textsc{No Recombination} \cite{jia2016data} & 85.2  & 58.1  & 78.0  & 71.4  & 76.4  & 79.6  & 76.2  & 81.4  & 75.8  \\
    \textsc{DataRecomb} \cite{jia2016data}*(+data) & 87.5  & 60.2  & 81.0  & 72.5  & 78.3  & 81.0  & 79.5  & 79.6  & 77.5  \\
    \textsc{CrossDomain} \cite{su2017cross} & \textbf{88.2} & 62.2 & \textbf{82.1} & \textbf{78.8} & 80.1 & \textbf{86.1} & 83.7 & 83.1 & \textbf{80.6} \\
    \textsc{DomainEncoding} \cite{herzig2017neural} & 86.2  & 62.7  & \textbf{82.1}  & 78.3  & {80.7}  & 82.9  & 82.2  & 81.7  & 79.6  \\
    \textsc{Seq2Action} \cite{chen2018sequence} & \textbf{88.2}  & 61.4  & 81.5  & 74.1  & {80.7}  & 82.9  & 80.7  & 82.1  & 79.0  \\
    \hline
    \textsc{Att}  & 86.2  & 61.4  & 76.4  & 68.8  & 75.2  & 76.9  & 78.9  & 82.2  & 75.7  \\
    \textsc{Att} + \textsc{Pseudo}($\mathcal{LF}$) & 87.2  & 60.9  & 73.2  & 71.4  & 75.8  & 80.1  & 79.2  & 82.0  & 76.2  \\
    \textsc{Att} + \textsc{Dual} & 87.5  & 63.7  & 79.8  & 73.0  & \textbf{81.4}  & 81.5  & 81.6  & 83.0  & 78.9  \\
    \textsc{Att} + \textsc{Dual} + $\mathcal{LF}$ & 88.0  & 65.2  & 80.7  & 76.7  & {80.7}  & 82.4  & \textbf{84.0}  & \textbf{83.8}  & 80.2  \\
    \hline
    \textsc{AttPtr} & 86.7  & 63.2  & 74.4  & 69.3  & 75.8  & 77.8  & 78.3  & 82.2  & 76.0  \\
    \textsc{AttPtr} + \textsc{Pseudo}($\mathcal{LF}$) & 85.7  & 63.4  & 74.4  & 69.8  & 78.3  & 78.7  & 79.8  & 82.1  & 76.5  \\
    \textsc{AttPtr} + \textsc{Dual} & 87.7  & 63.4  & 77.4  & 74.1  & 80.1  & 80.1  & 82.5  & 83.4  & 78.6  \\
    \textsc{AttPtr} + \textsc{Dual} + $\mathcal{LF}$ & 87.0  & \textbf{66.2}  & 79.8  & 75.1  & {80.7}  & 83.3  & 83.4  & \textbf{83.8}  & 79.9  \\
    \hline
    \end{tabular}
    \end{adjustbox}
  }
  \caption{Test accuracies on \textsc{Overnight} compared with previous systems.}
  \label{tab:stateOfArt}%
\end{table*}%

\begin{table}[htbp]
  \centering{
  \small
    \begin{tabular}{p{18em}c}
\hline   
\textbf{Method} & \multicolumn{1}{p{2em}}{\textbf{ATIS}} \\
    \hline
    \textsc{ZC07} \cite{zettlemoyer2007online} & 84.6  \\
    \textsc{FUBL} \cite{kwiatkowski2011lexical} & 82.8  \\
    \textsc{GUSP++} \cite{poon2013grounded} & 83.5  \\
    \textsc{TISP} \cite{zhao2014type} & 84.2  \\
    \hline
    \textsc{Seq2Tree} \cite{dong2016language} & 84.6  \\
    ASN+\textsc{SupAtt} \cite{rabinovich2017abstract} & 85.9  \\
    \textsc{Tranx} \cite{yin2018tranx} & 86.2  \\
    \textsc{Coarse2Fine} \cite{dong2018coarse} & 87.7  \\\hline
    \textsc{Att} & 84.4 \\
    \textsc{Att} + \textsc{Pseudo}($\mathcal{LF}$) & 83.3  \\
    \textsc{Att} + \textsc{Dual} & 86.4  \\
    \textsc{Att} + \textsc{Dual} + $\mathcal{LF}$ & 87.1  \\
    \hline
    \textsc{AttPtr} & 85.7 \\
    \textsc{AttPtr} + \textsc{Pseudo}($\mathcal{LF}$) & 86.2  \\
    \textsc{AttPtr} + \textsc{Dual} & 88.6  \\
    \textsc{AttPtr} + \textsc{Dual} + $\mathcal{LF}$ & \textbf{89.1}  \\
    \hline
    \end{tabular}%
  }
  \caption{Test accuracies on ATIS compared with previous systems.}
  \label{tab:atis}%
\end{table}%

The results are illustrated in Table \ref{tab:stateOfArt} and \ref{tab:atis}. \textsc{Att} and \textsc{AttPtr} represent that the primal/dual models are attention-based Seq2seq and attention-based Seq2seq with copy mechanism respectively.
We train models with the dual learning algorithm if \textsc{Dual} is included, otherwise we only train the primal model by supervised training. \textsc{$\mathcal{LF}$} refers to the synthesized logical forms. \textsc{Pseudo} uses the $LF2Q$ model and $\mathcal{LF}$ to generate pseudo-labeled data. From the overall results, we can see that:

1) Even without the additional logical forms by synthesizing, the dual learning based semantic parser can outperform our baselines with supervised training, e.g., ``\textsc{Att} + \textsc{Dual}'' gets much better performances than ``\textsc{Att} + \textsc{Pseudo}($\mathcal{LF}$)'' in Table \ref{tab:stateOfArt} and \ref{tab:atis}. We think the $Q2LF$ and $LF2Q$ models can teach each other in dual learning: one model sends informative signals to help regularize the other model. Actually, it can also be explained as a data augmentation procedure, e.g., $Q2LF$ can generate samples utilized by $LF2Q$ and vice versa. While the \textsc{Pseudo} greatly depends on the quality of pseudo-samples even if a discount factor is considered.

2) By involving the synthesized logical forms $\mathcal{LF}$ in the dual learning for each domain respectively, the performances are improved further. We achieve state-of-the-art performance (89.1\%)\footnote{Although there is another published result that achieved better performance by using word class information from Wiktionary \cite{wang-etal-2014-morpho}, it is unfair to compare it with our results and other previous systems which only exploit data resources of ATIS.} on ATIS as shown in Table \ref{tab:atis}. On \textsc{Overnight} dataset, we achieve a competitive performance on average (80.2\%). The best average accuracy is from \citet{su2017cross}, which benefits from cross-domain training. We believe our method could get more improvements with stronger primal models (e.g. with domain adaptation). Our method would be compatible with various models.


3) Copy mechanism can remarkably improve accuracy on ATIS, while not on \textsc{Overnight}. The average accuracy even decreases from 80.2\% to 79.9\% when using the copy mechanism. We argue that \textsc{Overnight} dataset contains a very small number of distinct entities that copy is not essential, and it contains less training samples than ATIS. This phenomenon also exists in \citet{jia2016data}.


\subsubsection{Ablation study}
\begin{table*}[htbp]
  \centering{
  \small
    \begin{tabular}{p{11.6em}|cccccccc|c|c}
    \hline
    \textbf{Method} & \multicolumn{1}{p{1.9em}}{\textbf{Bas.}} & \multicolumn{1}{p{1.9em}}{\textbf{Blo.}} & \multicolumn{1}{p{1.9em}}{\textbf{Cal.}} & \multicolumn{1}{p{1.9em}}{\textbf{Hou.}} & \multicolumn{1}{p{1.9em}}{\textbf{Pub.}} & \multicolumn{1}{p{1.9em}}{\textbf{Rec.}} & \multicolumn{1}{p{1.9em}}{\textbf{Res.}} & \multicolumn{1}{p{1.9em}|}{\textbf{Soc.}} & \multicolumn{1}{p{1.9em}|}{\textbf{Avg.}} & \multicolumn{1}{p{2em}}{\textbf{ATIS}} \\
    \hline
    \textsc{Att} & 80.1  & 55.4  & 61.9  & 53.4  & 60.2  & 64.4  & 71.1  & 76.8  & 65.4  & 78.6  \\
    \textsc{Att}+\textsc{Pseudo}($\mathcal{Q}$) & 80.1  & 59.4  & 60.1  & 52.9  & 59.6  & 66.2  & 73.8  & 79.0  & 66.4  & 78.8  \\
    \textsc{Att}+\textsc{Pseudo}($\mathcal{LF}$) & 83.9  & 60.4  & 64.3  & 54.5  & 58.4  & 69.0  & 70.5  & 77.3  & 67.3  & 77.9  \\
    \textsc{Att}+\textsc{Pseudo}($\mathcal{LF}$+$\mathcal{Q}$) & 80.6  & 59.1  & 61.9  & 57.1  & 62.7  & 65.3  & 73.2  & \textbf{79.8}  & 67.5  & 78.3  \\
    \textsc{Att}+\textsc{Dual} & 82.6  & 60.2  & 72.0  & 58.7  & 66.5  & 73.6  & 74.1  & 79.3  & 70.9  & 79.5  \\
    \textsc{Att}+\textsc{Dual}+$\mathcal{Q}$ & 83.9  & 60.7  & 70.2  & \textbf{60.8}  & 69.6  & 71.3  & \textbf{76.2}  & \textbf{79.8}  & 71.6  & 79.7  \\
    \textsc{Att}+\textsc{Dual}+$\mathcal{LF}$ & 83.4  & 61.4  & 71.4  & 59.3  & 70.2  & 73.1  & 75.3  & 78.6  & 71.6  & 80.4  \\
    \textsc{Att}+\textsc{Dual}+$\mathcal{LF}$+$\mathcal{Q}$ & \textbf{85.4}  & \textbf{62.9}  & \textbf{73.2}  & 59.3  & \textbf{72.0}  & \textbf{75.5}  & 75.6  & 79.5  & \textbf{72.9}  & \textbf{81.7} \\
    \hline
    \textsc{AttPtr} & 81.1  & 58.1  & 63.1  & 48.7  & 55.3  & 69.4  & 68.4  & 77.4  & 65.2  & 84.8  \\
    \textsc{AttPtr}+\textsc{Pseudo}($\mathcal{Q}$) & 82.1  & 59.6  & 61.3  & 47.6  & 57.1  & 72.2  & 69.9  & 78.4  & 66.0  & 85.0  \\
    \textsc{AttPtr}+\textsc{Pseudo}($\mathcal{LF}$) & 82.4  & 59.1  & 62.5  & 54.5  & 63.4  & 71.3  & 69.6  & 77.6  & 67.5  & 86.2  \\
    \textsc{AttPtr}+\textsc{Pseudo}($\mathcal{LF}$+$\mathcal{Q}$) & 81.3  & 59.4  & 65.5  & 49.2  & 58.4  & 72.7  & 72.0  & 78.6  & 67.1  & 85.0  \\
    \textsc{AttPtr}+\textsc{Dual} & 81.8  & 60.2  & 68.5  & 57.1  & 65.2  & 72.2  & 74.1  & 79.0  & 69.8  & 86.2  \\
    \textsc{AttPtr}+\textsc{Dual}+$\mathcal{Q}$ & 81.6  & 60.7  & 69.6  & 61.4  & 68.9  & 74.1  & \textbf{79.8}  & 80.1  & 72.0  & 86.6  \\
    \textsc{AttPtr}+\textsc{Dual}+$\mathcal{LF}$ & 82.6  & \textbf{62.2}  & 68.5  & \textbf{62.4}  & 69.6  & 73.1  & 77.4  & 79.4  & 71.9  & \textbf{87.3}  \\
    \textsc{AttPtr}+\textsc{Dual}+$\mathcal{LF}$+$\mathcal{Q}$ & \textbf{83.6}  & \textbf{62.2}  & \textbf{72.6}  & 61.9  & \textbf{71.4}  & \textbf{75.0}  & 76.5  & \textbf{80.4}  & \textbf{73.0}  & 86.8  \\
    \hline
    \end{tabular}%
  }
  \caption{Semi-supervised learning experiments. We keep 50\% of the training set as labeled data randomly, and leave the rest as unpaired queries($\mathcal{Q}$) and logical forms($\mathcal{LF}$) to simulate unsupervised dataset.} 
  \label{tab:ablation}%
\end{table*}%

\noindent\textbf{Semi-supervised learning} 
We keep a part of the training set as labeled data $\mathcal{T}$ randomly and leave the rest as unpaired queries ($\mathcal{Q}$) and logical forms ($\mathcal{LF}$) to validate our method in a semi-supervised setting. The ratio of labeled data is 50\%. \textsc{Pseudo} here uses the $Q2LF$ model and $\mathcal{Q}$ to generate pseudo-labeled data, as well as $LF2Q$ model and $\mathcal{LF}$. From Table \ref{tab:ablation}, we can see that the dual learning method outperforms the \textsc{Pseudo} baseline in two datasets dramatically. The dual learning method is more efficient to exploit unlabeled data. In general, both unpaired queries and logical forms could boost the performance of semantic parsers with dual learning.
\vspace{0.5em}

\begin{figure}[]
    \centering
    \includegraphics[width=0.45\textwidth]{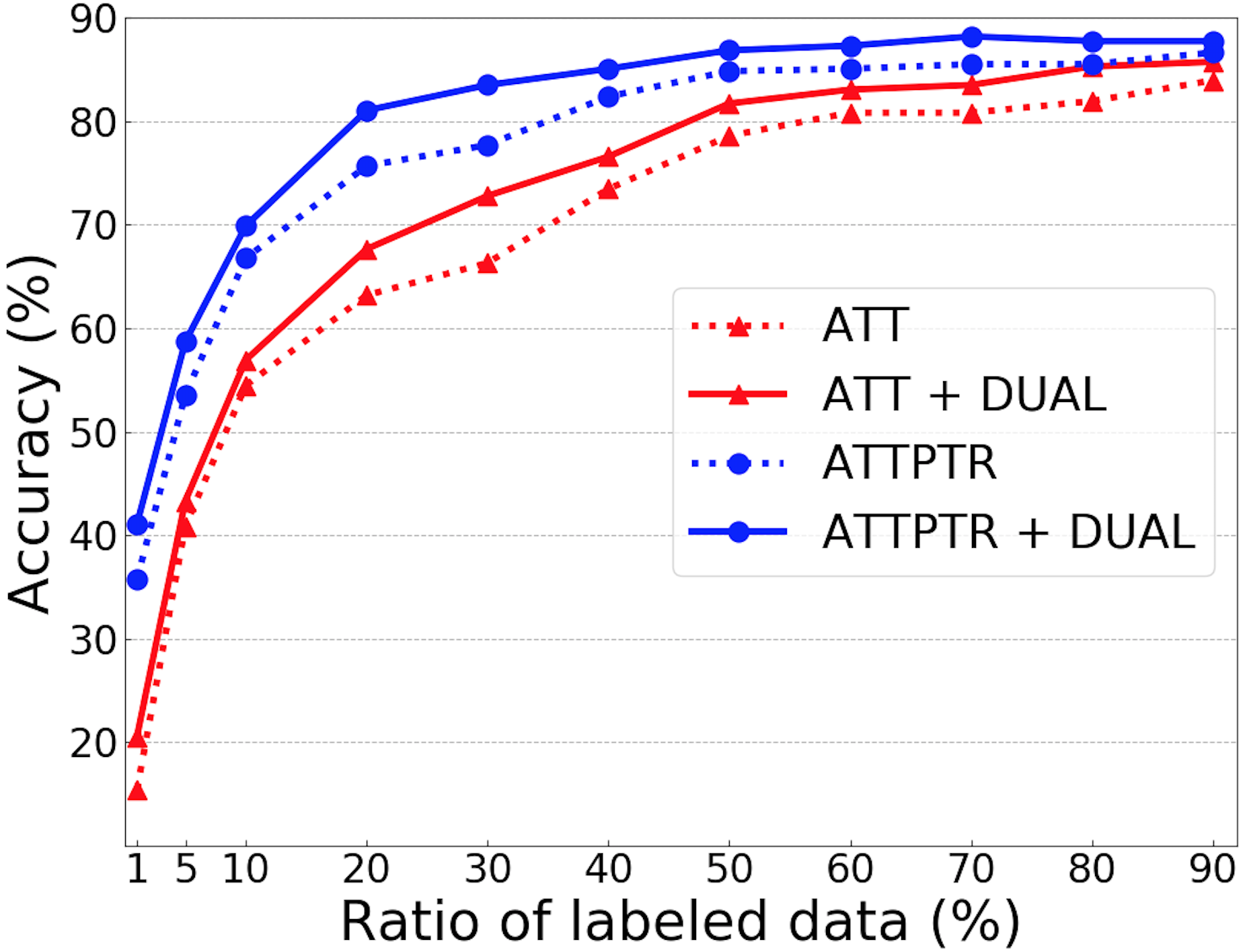}
    \caption{Test accuracies on ATIS. It varies the ratio of labeled data, and keeps the rest as unlabeled data.}
    \label{fig:ratio}
\end{figure}

\begin{figure}[htbp]
    \centering
    \includegraphics[width=0.45\textwidth]{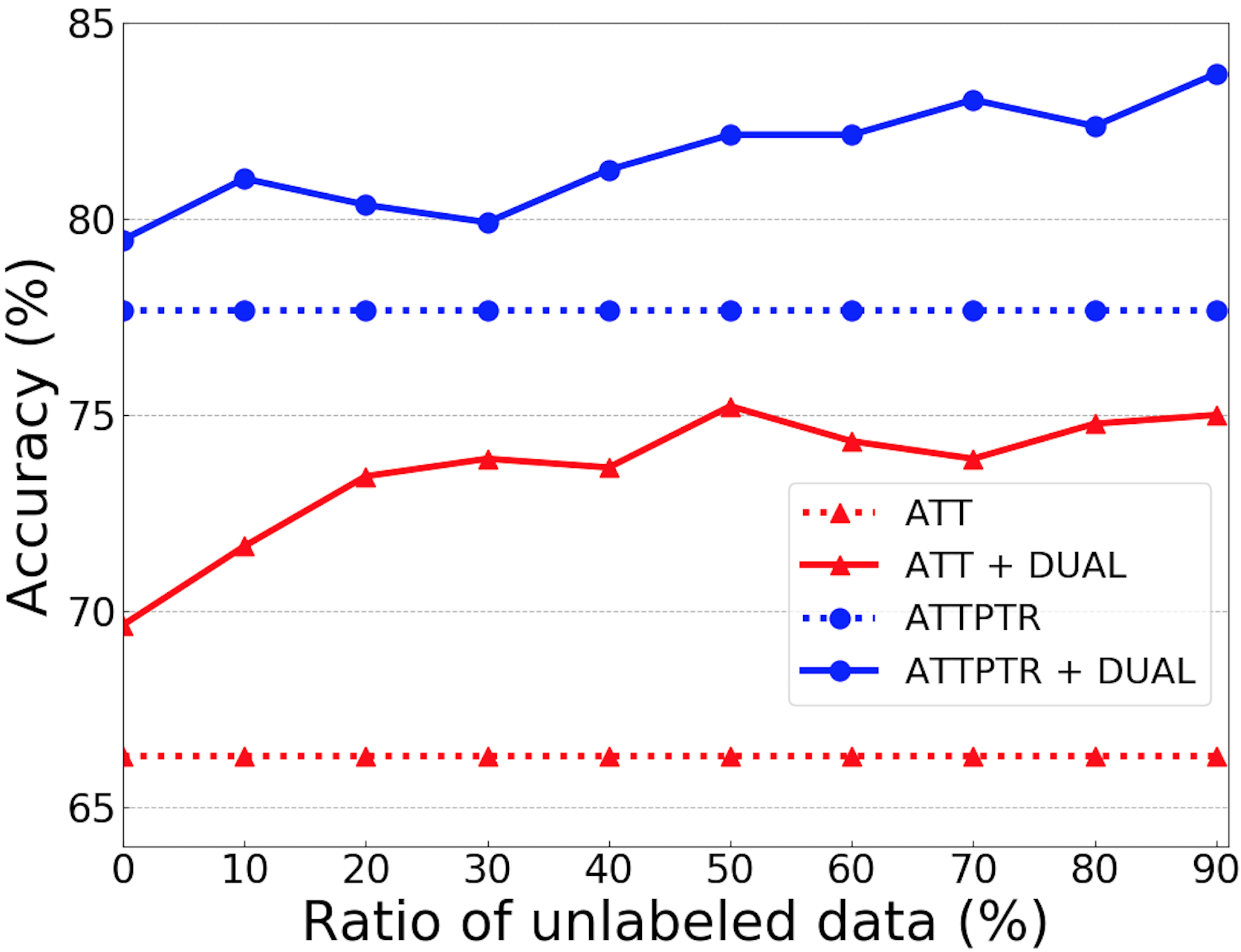}
    \caption{Test accuracies on ATIS. It fixes the ratio of labeled data as 30\%, and varies the ratio of unlabeled samples to the rest data.}
    \label{fig:unsupervised_ratio}
\end{figure}

\noindent\textbf{Different ratios} To investigate the efficiency of our method in semi-supervised learning, we vary the ratio of labeled data kept on ATIS from 1\% to 90\%. In Figure \ref{fig:ratio}, we can see that dual learning strategy enhances semantic parsing over all proportions. The prominent gap happens when the ratio is between 0.2 and 0.4. Generally, the more unlabeled data we have, the more remarkable the leap is. However, if the labeled data is really limited, less supervision can be exploited to keep the primal and dual models reasonable. For example, when the ratio of labeled data is from only 1\% to 10\%, the improvement is not that obvious. 

\noindent\textbf{Does more unlabeled data give better result?} We also fix the ratio of labeled data as 30\%, and change the ratio of unlabeled samples to the rest data on ATIS, as illustrated in Figure \ref{fig:unsupervised_ratio}. Even without unlabeled data (i.e. the ratio of unlabeled data is zero), the dual learning based semantic parser can outperform our baselines. However, the performance of our method doesn't improve constantly, when the amount of unlabeled data is increased. We think the power of the primal and dual models is constrained by the limited amount of labeled data. When some complex queries or logical forms come, the two models may converge to an equilibrium where the intermediate value loses some implicit semantic information, but the rewards are high.


\vspace{0.5em}

\begin{table}[htbp]
  \centering{
  \small
    \begin{tabular}{|l|c|c|c|}
    \hline
     Method & Validity & ATIS & \textsc{Overnight} \\
    \hline
    \textsc{Att} & $LM_{lf}$ & 80.6  & 71.5  \\
    \cline{2-4}
    \textsc{\ \ \ + Dual} & grammar check & 81.7  & 72.9  \\
    \hline
    \textsc{AttPtr} & $LM_{lf}$ & 86.2  & 71.4 \\
    \cline{2-4}
    \textsc{\ \ \ + Dual} & grammar check & 86.8  & 73.0 \\
    \hline
    \end{tabular}%
  }
  \caption{Test accuracies on ATIS and \textsc{Overnight} in semi-supervised learning setting (the ratio of labeled data is 50\%). On \textsc{Overnight}, we average across all eight domains. $LM_{lf}$ means using a logical form language model for validity reward, while ``grammar check'' means using the surface and semantic check. }
  \label{tab:lflm}%
\end{table}%

\noindent\textbf{Choice for validity reward} We conduct another experiment by changing validity reward in Eq.\ref{eq:val} with length-normalized LM score (i.e. language model of logical forms) like Eq.\ref{eq:lm}. Results (Table \ref{tab:lflm}) show that ``hard'' surface/semantic check is more suitable than ``soft'' probability of logical form LM. We think that simple language models may suffer from long-dependency and data imbalance issues, and it is hard to capture inner structures of logical forms from a sequential model. 

\section{Related Work}

\noindent\textbf{Lack of data}\quad A semantic parser can be trained from labeled logical forms or weakly supervised samples \cite{krishnamurthy2012weakly,berant2013semantic,liang2016neural,goldman2017weakly}. 
\citet{yih2016value} demonstrate logical forms can be collected efficiently and more useful than merely answers to queries. \citet{wang2015building} construct a semantic parsing dataset starting from grammar rules to crowdsourcing paraphrase. \citet{jia2016data} induces synchronous context-free grammar (SCFG) and creates new ``recombinant'' examples accordingly. \citet{su2017cross} use multiple source domains to reduce the cost of collecting data for the target domain. 
\citet{guo2018question} pre-train a question generation model to produce pseudo-labeled data as a supplement. In this paper, we introduce the dual learning to make full use of data (both labeled and unlabeled).
\citet{yin-etal-2018-structvae} introduce a variational auto-encoding model for semi-supervised semantic parsing.
Beyond semantic parsing, the semi-supervised and adaptive learnings are also typical in natural language understanding~\cite{Tur2005Combining,Bapna2017towards,Zhu2014Semantic,sz128-zhu-icassp18,sz128-zhu-sigdial18}.


\noindent\textbf{Constrained decoding}\quad To avoid invalid parses, additional restrictions must be considered in the decoding. 
\citet{dong2016language} propose \textsc{Seq2Tree} method to ensure the matching of parentheses, which can generate syntactically valid output.
\citet{cheng2017learning} and \citet{dong2018coarse} both try to decode in two steps, from a coarse rough sketch to a finer structure hierarchically.
\citet{krishnamurthy2017neural} define a grammar of production rules such that only well-typed logical forms can be generated.
\citet{yin2017syntactic} and \citet{chen2018sequence} both transform the generation of logical forms into query graph construction. 
\citet{zhao2019hierarchical} propose a hierarchical parsing model following the structure of semantic representations, which is predefined by domain developers.
We introduce a validity reward at the surface and semantic levels in the dual learning algorithm as a constraint signal.

\noindent\textbf{Dual learning}\quad Dual learning framework is first proposed to improve neural machine translation (NMT) \cite{he2016dual}. Actually, the primal and dual tasks are symmetric in NMT, while not in semantic parsing.
The idea of dual learning has been applied in various tasks \cite{xia2017dual}, such as Question Answering/Generation \cite{tang2017question,tang2018learning}, Image-to-Image Translation \cite{yi2017dualgan} and Open-domain Information Extraction/Narration \cite{sun2018logician}. We are the first to introduce dual learning in semantic parsing to the best of our knowledge.


\section{Conclusions and Future Work}
In this paper, we develop a semantic parsing framework based on dual learning algorithm, which enables a semantic parser to fully utilize labeled and even unlabeled data through a dual-learning game between the primal and dual models. We also propose a novel reward function at the surface and semantic levels by utilizing the prior-knowledge of logical form structures. Thus, the primal model tends to generate complete and reasonable semantic representation. Experimental results show that semantic parsing based on dual learning improves performance across datasets.

In the future, we want to incorporate this framework with much refined primal and dual models, and design more informative reward signals to make the training more efficient. It would be appealing to apply graph neural networks \cite{chen-etal-2018-structured,chen2019agentgraph} to model structured logical forms.

\section*{Acknowledgments}

This work has been supported by the National Key Research and Development Program of China (Grant No.2017YFB1002102) and the China NSFC projects (No. 61573241).


\bibliography{acl2019}
\bibliographystyle{acl_natbib}

\clearpage
\appendix
\onecolumn

\section{Detailed algorithm}
\label{app:dual_alg}
\begin{minipage}{\textwidth}
\begin{algorithm}[H]
\caption{Dual Learning Framework for Semantic Parsing}
\label{alg:dual_learning}
\begin{algorithmic}[1]
\Require \\
Supervised dataset $\mathcal{T}=\{\left<x,y\right>\}$; 
Unsupervised dataset for queries $\mathcal{Q}$ and logical forms $\mathcal{LF}$;\\
Pre-trained models on $\mathcal{T}$: $Q2LF$ model $P(y|x;\Theta_{Q2LF})$, $LF2Q$ model $P(x|y;\Theta_{LF2Q})$;\\
Pre-trained model on $\mathcal{Q}$ and queries of $\mathcal{T}$: Language Model for queries $LM_q$;\\
Indicator performs surface and semantic check for a logical form: $\text{grammar\_error\_indicator}(\cdot)$;\\
Beam search size $k$, hyper parameters $\alpha$ and $\beta$, learning rate $\eta_1$ for $Q2LF$ and $\eta_2$ for $LF2Q$;
\Ensure Parameters $\Theta_{Q2LF}$ of $Q2LF$ model
\Repeat
\State{
\Comment{\textbf{Reinforcement learning process uses unlabeled data, also reuses labeled data}}
\begin{multicols}{2}
\State Sample a query $x$ from $\mathcal{Q}\cup\mathcal{T}$;
\State $Q2LF$ model generates $k$ logical forms $y_1,y_2,\cdots,y_k$ via beam search;
\For{each possible logical form $y_i$}
\State Obtain validity reward for $y_i$ as $$R_q^{val}(y_i)=\text{grammar\_error\_indicator}(y_i)$$
\State Get reconstruction reward for $y_i$ as $$R_q^{rec}(x,y_i)=\log P(x|y_i;\Theta_{LF2Q})$$
\State Compute total reward for $y_i$ as $$r_i^q=\alpha R^{val}_q(y_i)+(1-\alpha)R^{rec}_q(x,y_i)$$
\EndFor
\State{Compute stochastic gradient of $\Theta_{Q2LF}$:
{\small
$$\nabla_{\Theta_{Q2LF}}\hat{E}[r]=\frac{1}{k}\sum_{i=1}^kr_i^q\nabla_{\Theta_{Q2LF}}\log P(y_i|x;\Theta_{Q2LF})$$}}
\State{Compute stochastic gradient of $\Theta_{LF2Q}$:
{\small
$$\nabla_{\Theta_{LF2Q}}\hat{E}[r]=\frac{1-\alpha}{k}\sum_{i=1}^k\nabla_{\Theta_{LF2Q}}\log P(x|y_i;\Theta_{LF2Q})$$}}
\State{Model updates:}
\begin{align*}
\Theta_{Q2LF}\leftarrow &\Theta_{Q2LF}+\eta_1\cdot\nabla_{\Theta_{Q2LF}}\hat{E}[r]\\
\Theta_{LF2Q}\leftarrow &\Theta_{LF2Q}+\eta_2\cdot\nabla_{\Theta_{LF2Q}}\hat{E}[r]
\end{align*}
\State{Sample a logical form $y$ from $\mathcal{LF}\cup\mathcal{T}$;}
\State{$LF2Q$ model generates $k$ queries $x_1,x_2,\cdots,x_k$ via beam search;}
\For{each possible query $x_i$}
\State{Obtain validity reward for $x_i$ as 
$$R_{lf}^{val}(x_i)=\log LM_q(x_i)/Length(x_i)$$}
\State{Get reconstruction reward for $x_i$ as
$$R_{lf}^{rec}(y,x_i)=\log P(y|x_i;\Theta_{Q2LF})$$}
\State{Compute total reward for $x_i$ as $$r_i^{lf}=\beta R^{val}_{lf}(x_i)+(1-\beta)R^{rec}_{lf}(y,x_i)$$}
\EndFor
\State{Compute stochastic gradient of $\Theta_{LF2Q}$:
{\small
$$\nabla_{\Theta_{LF2Q}}\hat{E}[r]=\frac{1}{k}\sum_{i=1}^kr_i^{lf}\nabla_{\Theta_{LF2Q}}\log P(x_i|y;\Theta_{LF2Q})$$}}
\State{Compute stochastic gradient of $\Theta_{Q2LF}$:
{\small
$$\nabla_{\Theta_{Q2LF}}\hat{E}[r]=\frac{1-\beta}{k}\sum_{i=1}^k\nabla_{\Theta_{Q2LF}}\log P(y|x_i;\Theta_{Q2LF})$$}}
\State{Model updates:}
\begin{align*}
\Theta_{LF2Q}\leftarrow &\Theta_{LF2Q}+\eta_2\cdot\nabla_{\Theta_{LF2Q}}\hat{E}[r]\\
\Theta_{Q2LF}\leftarrow &\Theta_{Q2LF}+\eta_1\cdot\nabla_{\Theta_{Q2LF}}\hat{E}[r]
\end{align*}
\end{multicols}}
\Comment{\textbf{After reinforcement learning process, use labeled data to fine-tune models}}
\State{Sample a $\left<x,y\right>$ pair from $\mathcal{T}$;}
\State Update $\Theta_{Q2LF}$ by $\Theta_{Q2LF}\leftarrow \Theta_{Q2LF}+\eta_1\cdot\nabla_{\Theta_{Q2LF}}\log P(y|x;\Theta_{Q2LF})$
\State Update $\Theta_{LF2Q}$ by $\Theta_{LF2Q}\leftarrow \Theta_{LF2Q}+\eta_2\cdot\nabla_{\Theta_{LF2Q}}\log P(x|y;\Theta_{LF2Q})$
\Until{$Q2LF$ model converges}
\end{algorithmic}
\end{algorithm}
\end{minipage}

\clearpage
\section{Examples of synthesized logical forms}
\label{app:examples_of_logical_form}
\begin{table}[H]
\centering
\small
\begin{tabular}{|p{20em}|p{20em}|}
\hline
\multicolumn{1}{|c|}{\textbf{Original}} & \multicolumn{1}{c|}{\textbf{After Modification}} \\
\hline
\multicolumn{2}{|c|}{\textbf{Entity Replacement}}\\
\hline
( lambda \$0 e ( and ( flight \$0 ) ( meal \$0 {\color{blue}\bf lunch:me} ) ( from \$0 ci0 ) ( to \$0 ci1 ) ) ) & ( lambda \$0 e ( and ( flight \$0 ) ( meal \$0 {\color{red}\bf dinner:me} ) ( from \$0 ci0 ) ( to \$0 ci1 ) ) )\\
\hline
( = al0 ( abbrev {\color{blue}\bf delta:al} ) ) & ( = al0 ( abbrev {\color{red}\bf usair:al} ) )\\
\hline
( lambda \$0 e ( and ( flight \$0 ) ( class\_type \$0 {\color{blue}\bf thrift:cl} ) ( from \$0 ci1 ) ( to \$0 ci0 ) ) ) & ( lambda \$0 e ( and ( flight \$0 ) ( class\_type \$0 {\color{red}\bf business:cl} ) ( from \$0 ci1 ) ( to \$0 ci0 ) ) )\\
\hline
\multicolumn{2}{|c|}{\textbf{Unary Replacement}}\\
\hline
( lambda \$0 e ( exists \$1 ( and ( {\color{blue}\bf round\_trip} \$1 ) ( from \$1 ci0 ) ( to \$1 ci1 ) ( = ( fare \$1 ) \$0 ) ) ) ) & ( lambda \$0 e ( exists \$1 ( and ( {\color{red}\bf oneway} \$1 ) ( from \$1 ci0 ) ( to \$1 ci1 ) ( = ( fare \$1 ) \$0 ) ) ) )\\
\hline
( lambda \$0 e ( and (  {\color{blue}\bf ground\_transport} \$0 ) ( to\_city \$0 ci0 ) ) ) & ( lambda \$0 e ( and ( {\color{red}\bf has\_meal} \$0 ) ( to\_city \$0 ci0 ) ) ) \\
\hline
( lambda \$0 e ( and ( {\color{blue}\bf taxi} \$0 ) ( to\_city \$0 ci0 ) ( from\_airport \$0 ap0 ) ) ) & ( lambda \$0 e ( and ( {\color{red}\bf limousine} \$0 ) ( to\_city \$0 ci0 ) ( from\_airport \$0 ap0 ) ) )\\
\hline
\multicolumn{2}{|c|}{\textbf{Binary Replacement}}\\
\hline
( lambda \$0 e ( and ( flight \$0 ) ( airline \$0 al0 ) ( {\color{blue}\bf approx\_departure\_time} \$0 ti0 ) ( from \$0 ci0 ) ( to \$0 ci1 ) ) ) & ( lambda \$0 e ( and ( flight \$0 ) ( airline \$0 al0 ) ( {\color{red}\bf approx\_arrival\_time} \$0 ti0 ) ( from \$0 ci0 ) ( to \$0 ci1 ) ) )\\
\hline
( lambda \$0 e ( and ( flight \$0 ) ( from \$0 ci0 ) ( to \$0 ci1 ) ( {\color{blue}\bf day\_return} \$0 da0 ) ( {\color{blue}\bf day\_number\_return} \$0 dn0 ) ( {\color{blue}\bf month\_return} \$0 mn0 ) ) ) & ( lambda \$0 e ( and ( flight \$0 ) ( from \$0 ci0 ) ( to \$0 ci1 ) ( {\color{red}\bf day\_arrival} \$0 da0 ) ( {\color{red}\bf day\_number\_arrival} \$0 dn0 ) ( {\color{red}\bf month\_arrival} \$0 mn0 ) ) )\\
\hline
( lambda \$0 e ( and ( flight \$0 ) ( airline \$0 al0 ) ( {\color{blue}\bf stop} \$0 ci0 ) ) ) & ( lambda \$0 e ( and ( flight \$0 ) ( airline \$0 al0 ) ( {\color{red}\bf from} \$0 ci0 ) ) )\\
\hline
\end{tabular}
\caption{Examples of synthesized logical forms on ATIS.}
\label{tab:atis_augmentation}%
\end{table}%
\begin{table}[htbp]
 \centering
 \small
 \begin{adjustbox}{width=\textwidth}
 \begin{tabular}{|c|c|m{0.8\textwidth}|}
  \hline
  \textbf{Domain} & \multicolumn{2}{c|}{\textbf{Logical Forms}}\\
  \hline
  \multirow{2}*{\textbf{ Bas.}} & pre & ( call SW.listValue ( call SW.getProperty ( ( lambda s ( call SW.filter ( var s ) ( string position ) ( string ! = ) {\color{blue}\bf en.position.point\_guard} ) ) ( call SW.domain ( string player ) ) ) ( string player ) ) ) \\
  \cline{2-3} 
  & new & ( call SW.listValue ( call SW.getProperty ( ( lambda s ( call SW.filter ( var s ) ( string position ) ( string ! = ) {\color{red}\bf en.position.forward} ) ) ( call SW.domain ( string player ) ) ) ( string player ) ) ) \\
  \hline
  \multirow{2}*{\textbf{Blo.}} & pre & ( call SW.listValue ( call SW.filter ( call SW.getProperty ( call SW.singleton en.block ) ( string ! type ) ) ( string shape ) ( string = ) {\color{blue}\bf en.shape.pyramid} ) ) \\ 
  \cline{2-3} 
  & new & ( call SW.listValue ( call SW.filter ( call SW.getProperty ( call SW.singleton en.block ) ( string ! type ) ) ( string shape ) ( string = ) {\color{red}\bf en.shape.cube} ) ) \\
  \hline
  \multirow{2}*{\textbf{Cal.}} & pre & ( call SW.listValue ( call SW.filter ( call SW.getProperty ( call SW.singleton en.location ) ( string ! type ) ) ( call SW.reverse ( string location ) ) ( string = ) {\color{blue}\bf en.meeting.weekly\_standup} ) ) \\ 
  \cline{2-3} 
  & new & ( call SW.listValue ( call SW.filter ( call SW.getProperty ( call SW.singleton en.location ) ( string ! type ) ) ( call SW.reverse ( string location ) ) ( string = ) {\color{red}\bf en.meeting.annual\_review} ) ) \\
  \hline
  \multirow{2}*{\textbf{Hou.}} & pre & ( call SW.listValue ( call SW.filter ( call SW.getProperty ( call SW.singleton en.housing\_unit ) ( string ! type ) ) ( string housing\_type ) ( string = ) ( call SW.concat {\color{blue}\bf en.housing.apartment en.housing.condo} ) ) ) \\ 
  \cline{2-3} 
  & new & ( call SW.listValue ( call SW.filter ( call SW.getProperty ( call SW.singleton en.housing\_unit ) ( string ! type ) ) ( string housing\_type ) ( string = ) ( call SW.concat {\color{red}\bf en.housing.condo en.housing.apartment} ) ) ) \\
  \hline
  \multirow{2}*{\textbf{Pub.}} & pre & ( call SW.listValue ( call SW.filter ( call SW.getProperty ( call SW.singleton en.article ) ( string ! type ) ) ( string author ) ( string = ) {\color{blue}\bf en.person.efron} ) ) \\ 
  \cline{2-3} 
  & new & ( call SW.listValue ( call SW.filter ( call SW.getProperty ( call SW.singleton en.article ) ( string ! type ) ) ( string author ) ( string = ) {\color{red}\bf en.person.lakoff} ) ) \\
  \hline
  \multirow{2}*{\textbf{Rec.}} & pre & ( call SW.listValue ( call SW.getProperty {\color{blue}\bf en.recipe.rice\_pudding} ( string cuisine ) ) ) \\ 
  \cline{2-3} 
  & new & ( call SW.listValue ( call SW.getProperty {\color{red}\bf en.recipe.quiche} ( string cuisine ) ) ) \\
  \hline
  \multirow{2}*{\textbf{Res.}} & pre & ( call SW.listValue ( call SW.getProperty {\color{blue}\bf en.restaurant.thai\_cafe} ( string neighborhood ) ) ) \\ 
  \cline{2-3} 
  & new & ( call SW.listValue ( call SW.getProperty {\color{red}\bf en.restaurant.pizzeria\_juno} ( string neighborhood ) ) ) \\
  \hline
  \multirow{2}*{\textbf{Soc.}} & pre & ( call SW.listValue ( call SW.getProperty ( ( lambda s ( call SW.filter ( var s ) ( string field\_of\_study ) ( string ! = ) {\color{blue}\bf en.field.computer\_science} ) ) ( call SW.domain ( string student ) ) ) ( string student ) ) ) \\ 
  \cline{2-3} 
  & new & ( call SW.listValue ( call SW.getProperty ( ( lambda s ( call SW.filter ( var s ) ( string field\_of\_study ) ( string ! = ) {\color{red}\bf en.field.history} ) ) ( call SW.domain ( string student ) ) ) ( string student ) ) ) \\
  \hline
 \end{tabular}
 \end{adjustbox}
 \caption{Examples of synthesized logical forms on \textsc{Overnight}.}
 \label{tab:overnight_augmentation}
\end{table}

\clearpage


\twocolumn

\end{document}